\documentclass{article}
\usepackage{ijcai20}

\usepackage{times}
\usepackage{soul}
\usepackage{url}
\usepackage[hidelinks]{hyperref}
\usepackage[utf8]{inputenc}
\usepackage[small]{caption}
\usepackage{graphicx}
\usepackage{amsmath}
\usepackage{amsthm}
\usepackage{booktabs}
\usepackage{algorithm}
\usepackage{algorithmic}
\usepackage{times}
\usepackage{epsfig}

\usepackage{amsmath}
\usepackage{amssymb}
\usepackage{comment}

\usepackage{booktabs}
\usepackage{multirow}
\usepackage{arydshln}
\usepackage{subfigure}
\usepackage{verbatim}

\title{Arbitrary Talking Face Generation via Attentional Audio-Visual Coherence Learning}

\author{
	Hao Zhu$^{1,2}$
	\and
	Huaibo Huang$^{2,3}$\and
	Yi Li$^{2,3}$\and
	Aihua Zheng$^{1}$\And
	Ran He$^{2,3}$\footnote{corresponding author} 
	\affiliations
	$^1$School of Computer Science and Technology, Anhui University, Hefei, China \\
	$^2$NLPR\&CEBSIT\&CRIPAC, Institute of Automation, CAS, Beijing, China\\
	$^3$School of Artificial Intelligence, University of Chinese Academy of Sciences, Beijing, China \\
	\emails
	haozhu96@gmail.com,
	\{huaibo.huang,yi.li\}@cripac.ia.ac.cn, 
	ahzheng214@ahu.edu.cn, 
	rhe@nlpr.ia.ac.cn
}

\begin{document}
	
	\maketitle
	
	\begin{abstract}
		Talking face generation aims to synthesize a face video with precise lip synchronization as well as a smooth transition of facial motion over the entire video via the given speech clip and facial image. Most existing methods mainly focus on either disentangling the information in a single image or learning temporal information between frames. However, cross-modality coherence between audio and video information has not been well addressed during synthesis. In this paper, we propose a novel arbitrary talking face generation framework by discovering the audio-visual coherence via the proposed Asymmetric Mutual Information Estimator (AMIE). In addition, we propose a Dynamic Attention (DA) block by selectively focusing the lip area of the input image during the training stage, to further enhance lip synchronization. Experimental results on benchmark LRW dataset and GRID dataset transcend the state-of-the-art methods on prevalent metrics with robust high-resolution synthesizing on gender and pose variations.
	\end{abstract}
	
	\section{Introduction}
	\label{sec:introduction}
	Talking face generation, which aims to generate a realistic talking video for the given still face image and speech clip, has been an active research topic. It has wide potential applications such as movie animation, teleconferencing, talking agents, and enhancing speech comprehension while preserving privacy. Although recent efforts have achieved impressive talking face synthesis for arbitrary identities, it is still a huge challenge due to the heterogeneous between audio and video, together with the appearance diversity of arbitrary identities.

		Existing state-of-the-art works \cite{Zhou2018TalkingFG,chen2019hierarchical} either try to disentangle the content and identity features from the speech for the video generation step or leverage the landmark as a middle feature to bridge the gap between audio and video. 
		Despite the great progress on feature representation for video generation, they usually leverage reconstruct loss between generated and real frames, while neglecting an essential problem in talking face generation: how to sufficiently express the audio feature into generated video? 
		To reduce the uncertainty in the audio-to-video generation process and ensure the synchronized talking face transition, we argue that the giving audio and generated video should share maximum information, i.e., the minimized sharing entropy across modalities. Herein, we propose audio-visual coherence learning, which learns the shared entropy between audio and visual modalities to generate talking face video with precise lips shape.
	
	\begin{figure}[tb]
		\begin{center}
			\includegraphics[width=0.85\linewidth]{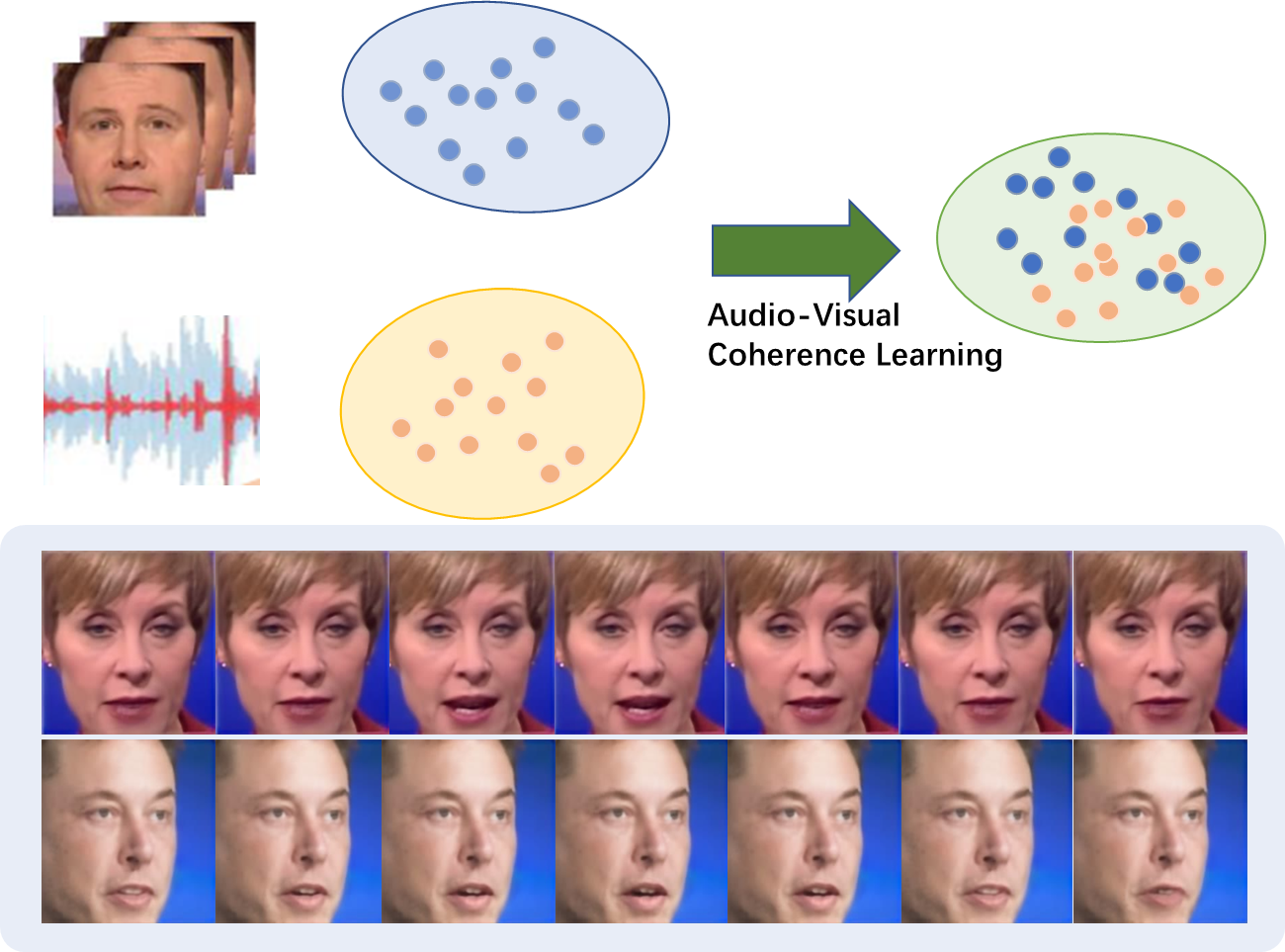}
		\end{center}
		\caption{Illustration of proposed audio-visual coherence learning.}
		\label{fig:mutual-demo}
	\end{figure}
	
	In the AI community, it is always active to exploit information theoretic measures to solve real-world problems \cite{he2009robust,jun2011compact,he2014robust,chen2016infogan}. 
	Mutual information (MI), as a commonly used measure to explore the coherence between two distributions, has been successfully applied to feature selection \cite{sotoca2010supervised,liu2009feature}, face analysis \cite{jun2011compact}, and gene expression \cite{gupta2010mib}. 
	As explained in \cite{vincent2010stacked}, mutual information can be utilized to learn a parametrized mapping from a given input to a higher-level representation while preserving information of the original input. This can be referred to as the infomax principle translating to maximize the mutual information between the output and input during the generative network. 
	For instance, InfoGAN \cite{chen2016infogan} maximized the MI between generated samples and latent codes to learn more interpretable representations during generation. 
	
	We argue that the audio and corresponding facial information contains potential mutual dependency, therefore, MI is able to calculate the shared information entropy between audio and video if we can maximize the MI between them, that means, we minimized the uncertainty in audio-to-video expression process, which can ensure more stable and precise lip motion generation. 
	Therefore, we propose to explore the cross-modal audio-visual coherence via MI in this paper. 
	
	Belghazi \textit{et al.} \shortcite{Belghazi2018MutualIN} recently presented a Mutual Information Neural Estimator (MINE) to explore a KL-based MI lower bound and train this estimator on target distributions. Unfortunately, it cannot be directly applied to talking face generation.
	On the one hand, if MINE for GANs uses generated sample and input pairs for both training and estimating, the mutual information estimation may be misled due to the low quality of generation in the early stages. Thinking that GANs push the generated sample distribution to the real frame distribution, we propose to use the real sample and its corresponding input to update the mutual information estimator, followed by asymmetrically maximizing the mutual information between the generated sample and input for better GAN training.
	On the other hand, our objective is to maximize the MI between heterogeneous modalities while not concerning the accurate MI value \cite{Belghazi2018MutualIN}, we therefore replace the original formulation of MI by Jensen-Shannon represented MI estimator, which estimates the relative amplitude of MI rather than the exact value, and has been well studied in neural network optimization with empirically more stable learning \cite{hjelm2018learning}. 
	We refer to our strategy as Asymmetric Mutual Information Estimator (AMIE).

	In addition, conventional approaches \cite{Chung2017YouST,Chen2018LipMG,Vougioukas2018EndtoEndSF,Zhou2018TalkingFG,chen2019hierarchical} directly employ the whole global area on given face image during synthesis, which is difficult for neural networks to discover the relation between audio and local lips. We observe that a talking face video is mainly composited by the identity-related feature and the lip-related feature, where the former is more stable while the latter is more temporal dynamic. Therefore, it is essential to separate these two features for arbitrary identity generation. Herein, we propose to leverage the given face and previous generated frame to provide identity-related and lip-related information, respectively. 
	We propose a dynamic attention block on lip area to preserve the identity information and leverage feature of lip motion, i.e., paying different attentions on given face image (identity-related) and the previous generated frame (lip-related) during different training stages. 
	
	\begin{figure*}[t]
		\begin{center}
			\includegraphics[width=0.78\linewidth]{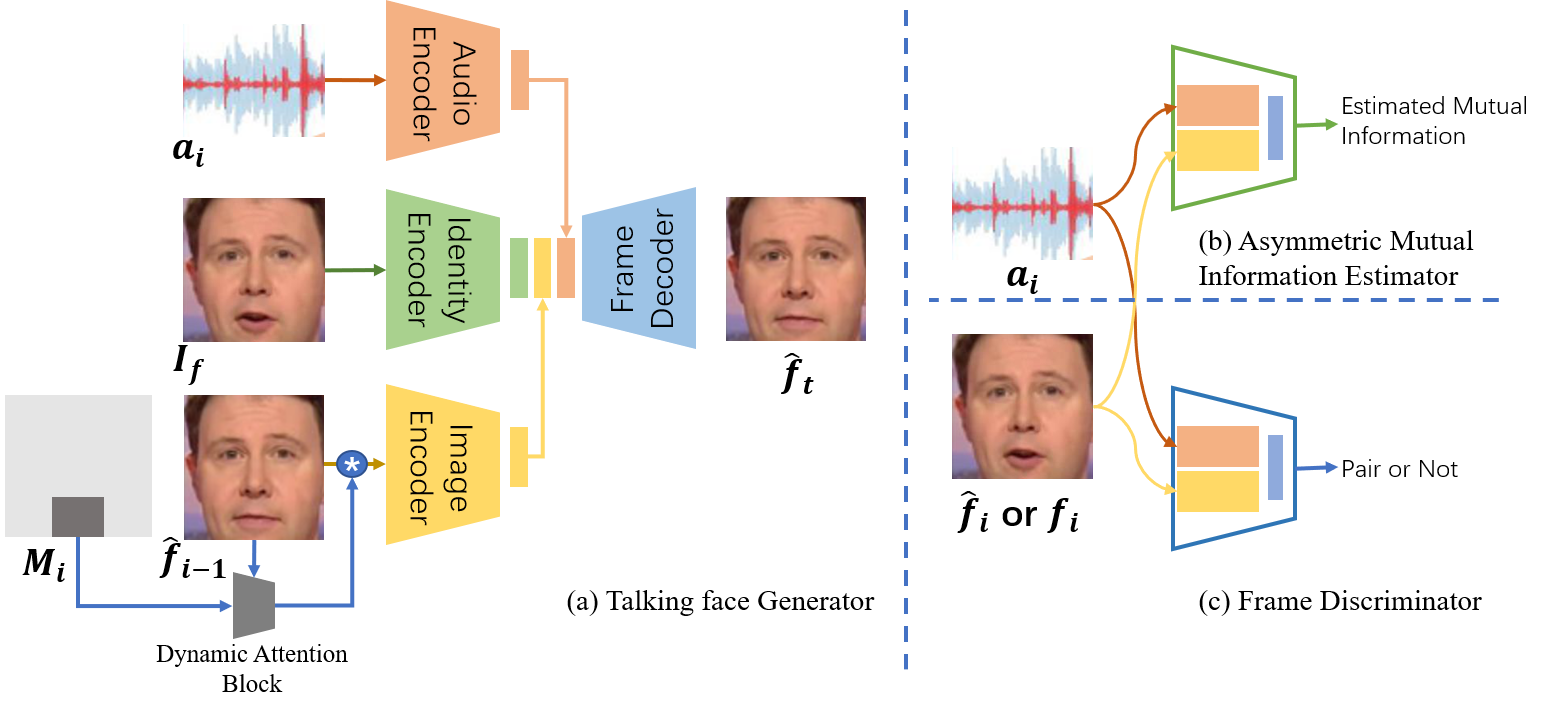}
		\end{center}
		\caption{Pipeline of our proposed method.}
		\label{fig:archi}
	\end{figure*}

\begin{figure}[b]
		\centering
		\includegraphics[width=0.88\linewidth]{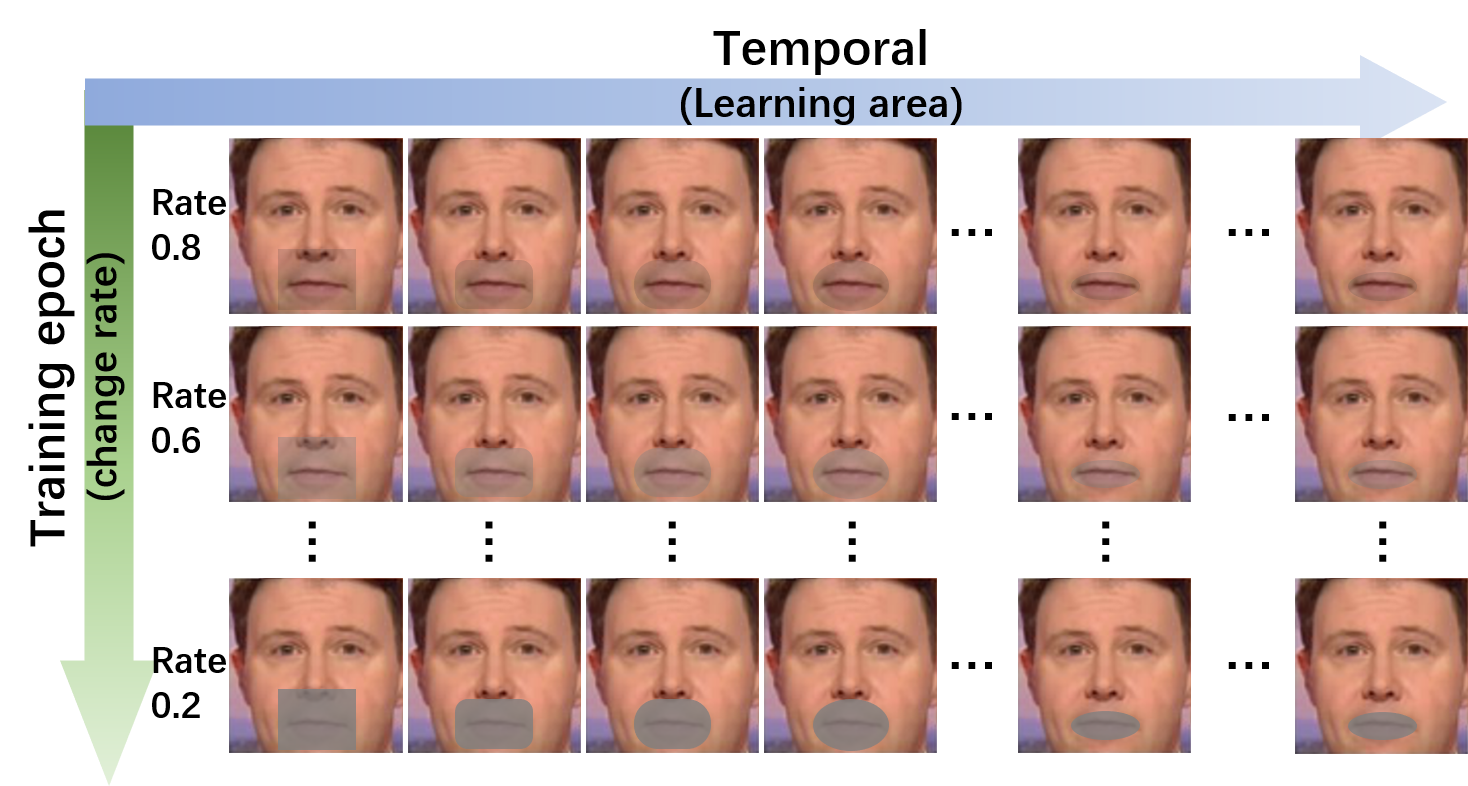}
		\caption{The illustration of the proposed dynamic attention. 
		}
		\label{fig:disentangle_demo}
	\end{figure}

	Based on the above discussion, we propose a novel and robust method by exploring the coherence between audio and visual modalities for arbitrary talking face generation in this paper.    
	The proposed model consists of three components: talking face generator, asymmetric mutual information estimator, and frame discriminator, as shown in Fig. \ref{fig:archi}.
	First, the talking face generator is designed to generate target frames from the given input: one audio clip, one still facial image, and the previous generated frame. 
	Then, the asymmetric mutual information estimator is introduced to maximize the mutual information between generated video and audio distributions via the information learned from a neural network based on MI measure. 
	Finally, we feed the generated frame and audio into frame discriminator to detect whether they are matched or not.
	In the training stage, the lip area will give different attention to make the network focus on the most important area by leveraging the Dynamic Attention block, as illustrated in Fig \ref{fig:disentangle_demo}.
	
	The main contributions of our paper can be summarized as follows:
	\begin{itemize}
		\item We propose to discover the audio-visual coherence via the Asymmetric Mutual Information Estimator (AMIE) to better express the audio information into generated video in talking face generation. 
		\item We design a dynamic attention block to improve the transition of generated video for arbitrary identities by decoupling the lip-related and identity-related information.
		\item Extensive experiments yield a new state-of-the-art on benchmark datasets LRW \cite{Chung16} and GRID \cite{cooke2006audio} with robust high-resolution synthesizing on gender and pose variations. 
	\end{itemize}

	\section{Related Works}
	\label{sec:related_work}
	In this section, we briefly review the related works on talking face generation and mutual information estimation.
	
	\subsection{Talking Face Generation}
	Earlier works on talking face generation mainly synthesized the specific identity from the dataset by given arbitrary speech audio.
	Rithesh \textit{et al.} \shortcite{kumar2017obamanet} used a time-delayed LSTM to generate key points synced to the audio and use another network to generate the video frames conditioned on the key points. Supasorn \textit{et al.} \shortcite{suwajanakorn2017synthesizing} proposed a teeth proxy to improve the quality of the teeth during generation.
	Next, some works tried to synthesized the talking faces for the identities from the dataset \cite{Chung2017YouST,Vougioukas2018EndtoEndSF,Chen2018LipMG}.
	Recently, the synthesis of the talking face for the arbitrary identities out of the dataset has drawn much attention. Zhou \textit{et al.} \shortcite{Zhou2018TalkingFG} proposed an adversarial learning method to disentangle the different information for one image during generation. 
	Chen \textit{et al.} \shortcite{Zhou2018TalkingFG} proposed to leverage landmark as middle information to better guide face generation. 
	However, they mainly focused on the inner-modal coherence, while lacking of discovering the cross-modal coherence. 
	\subsection{Mutual Information Estimation}
	It is historically difficult to measure the mutual dependence between two random variables in information theory. Some past solutions are only suitable for discrete variables or limited known probability distributions. 
	Previous works tried to employ parameters-free approaches \cite{kraskov2004estimating}, or relied on approximate Gaussianity of data distribution \cite{hulle2005edgeworth} to estimate the mutual information. 
	Recently, Belghazi \textit{et al.} \shortcite{Belghazi2018MutualIN} proposed a backpropagation MI estimator that exploited a dual optimization based on dual representations of the KL-divergence \cite{ruderman2012tighter} to estimate divergences beyond the minimax objective as formalized in GANs. 
	
	\section{Proposed Method}
	\label{sec:approaches}
	In this paper, we propose a novel model for arbitrary talking face generation by attentively discovering the cross-modal coherence, as shown in Fig. \ref{fig:archi}. 
	We first overview the architecture of our model, followed by the elaboration of two novel components: Asymmetric Mutual Information Estimator (AMIE) and Dynamic Attention (DA) modules. The training details are provided at the end of this section.

	\subsection{Overview}
	\label{overview1}
	Our model consists of three parts: a Talking Face Generator, a Frame Discriminator, and a Mutual Information Estimator. 
	
	\paragraph{Talking Face Generator.}
	There are three inputs of the generator: 1) the input face $I_f$ which ensures the texture information of the output frame, 2) the speech audio clip $A = \{a_1, a_2, ..., a_n\}$, working as the condition to supervise the lip changing, and 3) the previously generated frame $\hat{f}_{i-1} \in \hat{F}=\{\hat{f}_1,\hat{f}_2,...,\hat{f}_n\}$ where the i-th frame guarantees the smoothness of the image generation by feeding more temporal information. The three inputs are fed to the Identity Encoder, Audio Encoder, and Image Encoder respectively, while the target video frame $\hat{f}_i$ is generated by the Frame Decoder. The generated frames $\hat{F}$ should be similar to real frames $F = \{f_1,f_2,...,f_n\}$.

	\paragraph{Asymmetric Mutual Information Estimator (AMIE).}
	AMIE aims to approximate the mutual information between the generated video frame and audio. 
	It is trained using $\{f_i, a_i\}$ and $\{f_j, a_k\}$, where $j$ and $k$ are random sampled from $1\sim n$. These two pairs are serve as samples that the sampled from joint and marginal distributions.
	While in the estimating stage, we asymmetrically estimate mutual information using $\{\hat{f}_i, a_i\}$, and $\{\hat{f}_j, a_k\}$, i.e., we use the real pairs for training while the generated samples to estimate. 
	It consists of an Image Encoder and an Audio Encoder having the same architecture as defined in the generator, followed by a 3-layer classifier with the output as a  1-dimension scalar. We shall elaborate on the details of the proposed AMIE in the following section.
	
	\paragraph{Frame Discriminator.}
	Frame discriminator is fed by the pairs of the real frame and audio clip $\{f_i, a_i\}$, or the pairs of the generated frame and corresponding audio clip $\{\hat{f}_{i}, a_i\}$.
	The output of the discriminator is the probability of whether the inputs (audio and frame) are matched.
	The discriminator consists of an Image CNN, an Audio FC, and a classifier. 
	We flatten the output of Image CNN, and the Audio FC then feed concatenated features to the final classifier to produce 1-dimensional output. 
	
	\subsection{Asymmetric Mutual Information Estimator}
	\label{mia}
	We first introduce the theory of mutual information neural estimation (MINE), followed by the mutual information estimation in talking face generation task which leverages the proposed AMIE to learn the cross-modal coherence. 
	\subsubsection{Preliminary Theory of 'MINE'}
	Mutual information is a measurement of mutual dependency between two probability distributions, 
	\begin{equation}
	I(X,Y)=\sum_{y \in Y}\sum_{x\in{X}}p(x,y)log\frac{p(x,y)}{p(x)p(y)},
	\label{eq:definition_mi}
	\end{equation}
	where $p(x, y)$ is the joint probability function of $X$ and $Y$, and $p(x)$ and $p(y)$ are the marginal probability distribution functions of $X$ and $Y$ respectively.
	
	Clearly, mutual information is equivalent to the Kullback-Leibler (KL-) divergence between the joint $p(x, y)$ and the product of the marginal distributions $p(x)$ and $p(y)$:
	
	\begin{equation}
	I(X,Y) = D_{KL}(p(x,y)\parallel p(x)p(y)),
	\label{eq:definition_mi_kl}
	\end{equation}
	where $D_{KL}$ is defined as,
	\begin{equation}
	D_{KL}(p \parallel q) = \sum_{x \in X}p(x)log\frac{p(x)}{q(x)}.
	\end{equation}
	
	Furthermore, the $KL$ divergence admits the following  Donsker-Varadhan ($DV$) representation \cite{donsker1983asymptotic,Belghazi2018MutualIN} :
	\begin{equation}
	D_{KL}(p \parallel q) = \sup_{T:\omega\rightarrow\mathbb{R}} \mathbb{E}_p[T]-log(\mathbb{E}_q[e^T]),
	\label{eq:dv-representation}
	\end{equation}
	where the supremum is taken over all functions $T$ and $\omega \subset \mathbb{R}^d$ so that the two expectations are finite. Therefore, we leverage the bound:
	\begin{equation}
	I(X, Y)\geq I^{DV}_\Theta(X, Y), 
	\end{equation}
	where $I^{DV}_\Theta(X,Y)$ denotes the neural information measure, 
	\begin{equation}
	\begin{split}
	I^{DV}_\Theta(X, Y) = \sup_{\theta \in \Theta}\mathbb{E}_{p(x, y)}[T_\theta(x, y)]- \\
	log(\mathbb{E}_{p(x)p(y)}[e^{T_\theta(x, y)}] ),
	\end{split}
	\label{eq:final_mi}
	\end{equation}
	and $T_\theta$ denotes a neural network trained by maximizing the mutual information \cite{Belghazi2018MutualIN}. 
	
	\subsubsection{'AMIE' in Talking Face Generation} 
	In this cross-modal talking face generation task, we propose to explore the mutual information between the audio and the visual modality via an Asymmetric Mutual Information Estimator. 
	Given $\hat{F}$ and $A$ as the generated video frames and the given audios respectively, the neural network $T_\theta$ is fed by $\{\hat{f}_i,a_i\}$ and $\{\hat{f}_j,a_k\}$ to output a scalar which will be used to estimate the MI. 
	Furthermore, as we do not concern the accurate value of MI while maximizing it, inspired by \cite{hjelm2018learning}, we replace Donsker-Varadhan (DV) representation by Jensen-Shannon representation (JS), which has been noted with more stable learning in neural network optimization. 
	Therefore, the estimated mutual information between $\hat{F}$ and $A$ is: 
	\begin{equation}
	\begin{split}
	I^{JS}_\Theta(\hat{F}, A) = \sup_{\theta \in \Theta} 
	&\mathbb{E}_{p(\hat{f},a)}[-\varphi(-T_\theta(\hat{f}_i,a_i))]\;- \\
	&\mathbb{E}_{p(\hat{f})p(a)}[\varphi(T_\theta(\hat{f}_j,a_k))],
	\end{split}
	\end{equation}
	where $p(\hat{f}, a)$, $p(\hat{f})$ and $p(a)$ denote the joint distribution of the generated frame and audio, the marginal distributions of generated frame and the marginal distributions of generated audio respectively, in addition, $\varphi(\cdot)$ represents softplus operation:
	\begin{equation}
	\varphi(x)=log(1+e^x).
	\end{equation}

	\begin{figure}[t]
		
		\begin{center}
			\includegraphics[width=0.9\linewidth]{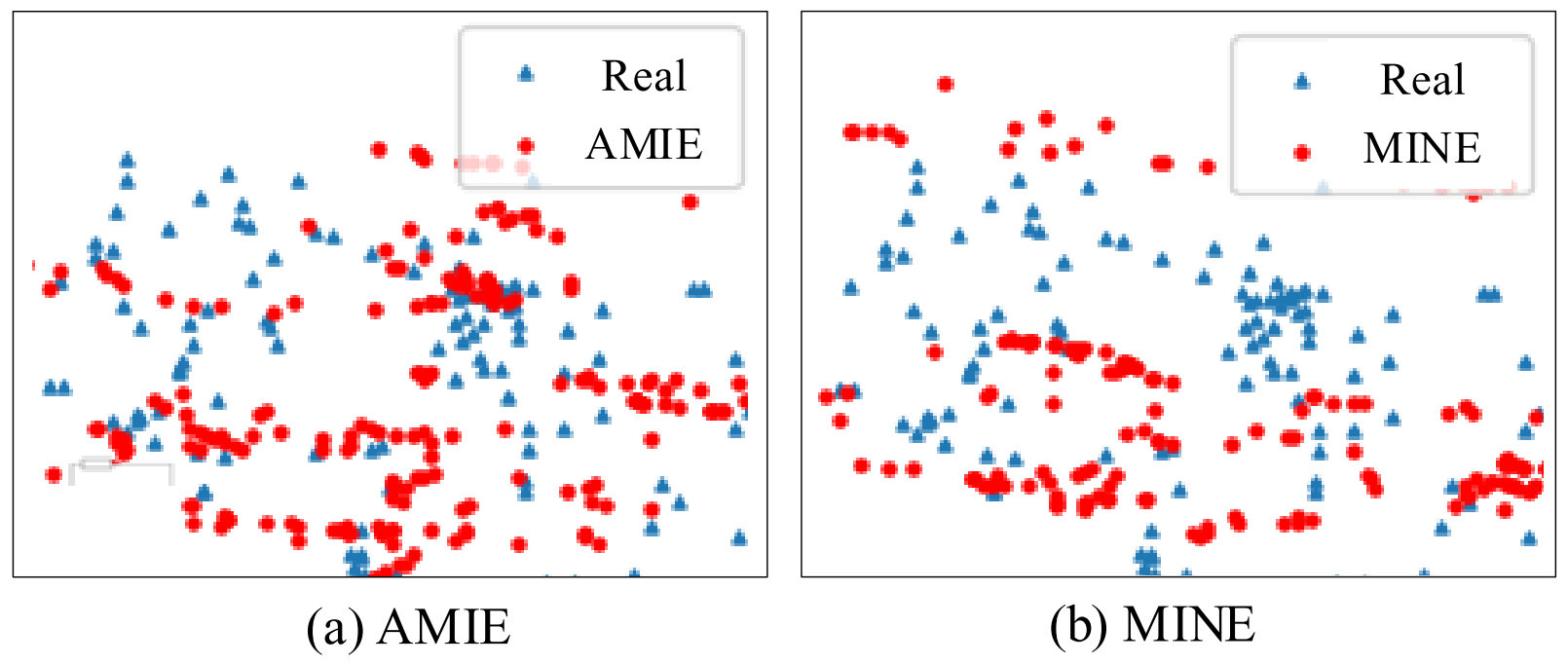}
		\end{center}
		\caption{Visualization of distributions of real and generated frames. We reduce the dimension of frames into two-dimension via PCA for better demonstration. It is obvious that the generated samples are closer to the real samples than that with original MINE. }
		\label{fig:distributions}
	\end{figure}
	
	As proved in \cite{Belghazi2018MutualIN}, one can estimate the mutual information between the generated video frame and audio by directly training $T_\theta$ on them. However, this may disturb the MI estimation since the generated frames are blurry and not accurate at earlier training epochs. 
	Furthermore, GANs are usually used to learn the probability distribution consistent with the real data, and mutual information is used to estimate the amount of shared information between the two distributions. 
	Therefore, our solution is to use mutual information in three distributions, we use the real frame and audio distribution during the training while the generated frame and audio distribution. 
	
	Specifically, let $F$ represent real video frames, we update the mutual information estimator by maximizing mutual information between the real frames and the audio
	($I^{JS}_\Theta(F, A)$),
	then use the updated estimator to maximize the mutual information between the generated frames and audio 
	($I^{JS}_\Theta(\hat{F}, A)$)
	for better GAN training. We refer to this as Asymmetric Mutual Information Estimator (AMIE) in this paper which can improve the quality of generation as verified in our experiments. As shown in Fig. \ref{fig:distributions}, comparing our AMIE with conventional MINE, we see that our generated samples are closer to the real samples than those from MINE.


	\subsection{Dynamic Attention on Lip Area}
	
	\label{da}
	
	As we discussed in Sec. \ref{sec:introduction}, we observe that a talking face video is mainly composited by the identity-related and lip-related features. Directly leveraging the whole area of the given face tends to generate face images with a slight jitter problem. 
	In order to capture different information among the given faces and disentangle the identity-related and lip-related information, we introduce a dynamic attention block. 
	When generating a video sequence, we assign an initial attention rate to the first frame, then predict the fine-grained attention mask for the following video frames. 
	%
	When given low attention on the lip, the input image mainly contains identity-related information (facial texture without the lip-related information), and the previous frame and audio clip together provide the information about the shape of the lip. Therefore, our model can divide the feature of one identity into two parts: identity-related feature and lip-related feature. The lower attention, the information is separated more completely. 
	
	However, it may significantly affect the quality of generation in the early stage if we directly assign a low initial attention mask $M_i$ on the lip since lacking supervision of lip information. Therefore, we progressively decrease the initial attention after several training epochs, which will enforce the visual information of the lip deriving from the previous frame $\hat{f}_{i-1}$. 
	Specifically, in the training stage, we start from relatively high attention (rate = $0.7\!\sim\!0.9$), and progressively decrease it to relatively low attention (rate = $0.1\!\sim\!0.3$), then we fix the rate to 1 for the last few epochs. 
	The attention masks of the following generated frame are predicted by the previous attention mask and the generated faces. 
	In practice, we give a coarse mask area and an attention rate, the following frames generation only predict a fine-grained mask area which is close to the lip shape, while leaving the rate unchanged. 
	
	
	
	%
	
	\subsection{Training Details}
	\label{train_details}
	In the training stage, 
	frame discriminator predicts the probability of whether frame and audio are paired or not, resulting in the following loss of our GAN:
	\begin{equation}
	\begin{split}
	\mathcal{L}_{GAN}(D,G) \;= \;&\mathbb{E}_{f\sim{P_d}}[log(D(f_i, a_i))]\,+ \\
	&\mathbb{E}_{z\sim{P_z}}[log(1-D(G(a_i, z),a_i))].
	\end{split}
	\label{eq:loss_gan}
	\end{equation}
	
	In order to synchronize the lip movements more accurately, we employ perceptual loss \cite{johnson2016perceptual} to capture high-level features differences between generated images and ground-truth. The perceptual loss is defined as:
	
	\begin{equation}
	\mathcal{L}_{perc}(f_i,\hat{f}_i) = \parallel \phi(f_i) - \phi(\hat{f}_i) \parallel_2^2,
	\label{eq:loss_prec}
	\end{equation}
	where $\phi$ is a feature extraction network.
	
	To focus on the lip movement, we only utilize the lip area of the frame for $L_1$ reconstruction loss,
	\begin{equation}
	\mathcal{L}_{lip}(f_i,\hat{f}_i)   = \parallel corp(f_i)-corp(\hat{f}_i)  \parallel_1,
	\label{eq:loss_mouth}
	\end{equation}
	where $corp(\cdot)$ means the cropped lip area from image. 
	
	While $\mathcal{L}_{perc}$ and $\mathcal{L}_{lip}$ measure the distance between visual concept, it is also important to shorten the distance between audio and visual modalities in high-level representation. 
		
	We implement mutual information as described before. We try to maximize it between generated frames and audios, 
	\begin{equation}
	\mathcal{L}_{mi}(\hat{F}, A)   = -I^{JS}_\Theta(\hat{F}, A).
	\label{eq:loss_mi}
	\end{equation}
	
	Our full model is optimized according to the following objective function:
	\begin{equation}
	\mathcal{L} = \mathcal{L}_{GAN} + \lambda_1\mathcal{L}_{perc} + \lambda_2\mathcal{L}_{lip} + \lambda_3\mathcal{L}_{mi}.
	\label{eq:loss_target}
	\end{equation}

	\section{Experiments}
	\label{sec:experiments}
	\subsection{Dataset and Metrics}
	
	We evaluate our method on prevalent benchmark datasets LRW \cite{Chung16} and GRID \cite{cooke2006audio}. 
	Frames are aligned into $256 \times 256$ faces and audios are processed into (Mel Frequency Cepstrum Coefficient) MFCC features at the sampling rate of 5000Hz. Then we match each frame with an MFCC audio input with a size of $20 \times 13$.
	We use common reconstruction metrics such as PSNR and SSIM \cite{wang2004image} to evaluate the quality of the synthesized talking faces. 
	Furthermore, we use Landmark Distance (LMD) to evaluate the accuracy of the generated lip by calculating the landmark distance between the generated video and the original video. The lower LMD, the better of the generation.

	\begin{figure}[tb]
		\centering
		\includegraphics[width=0.95\linewidth]{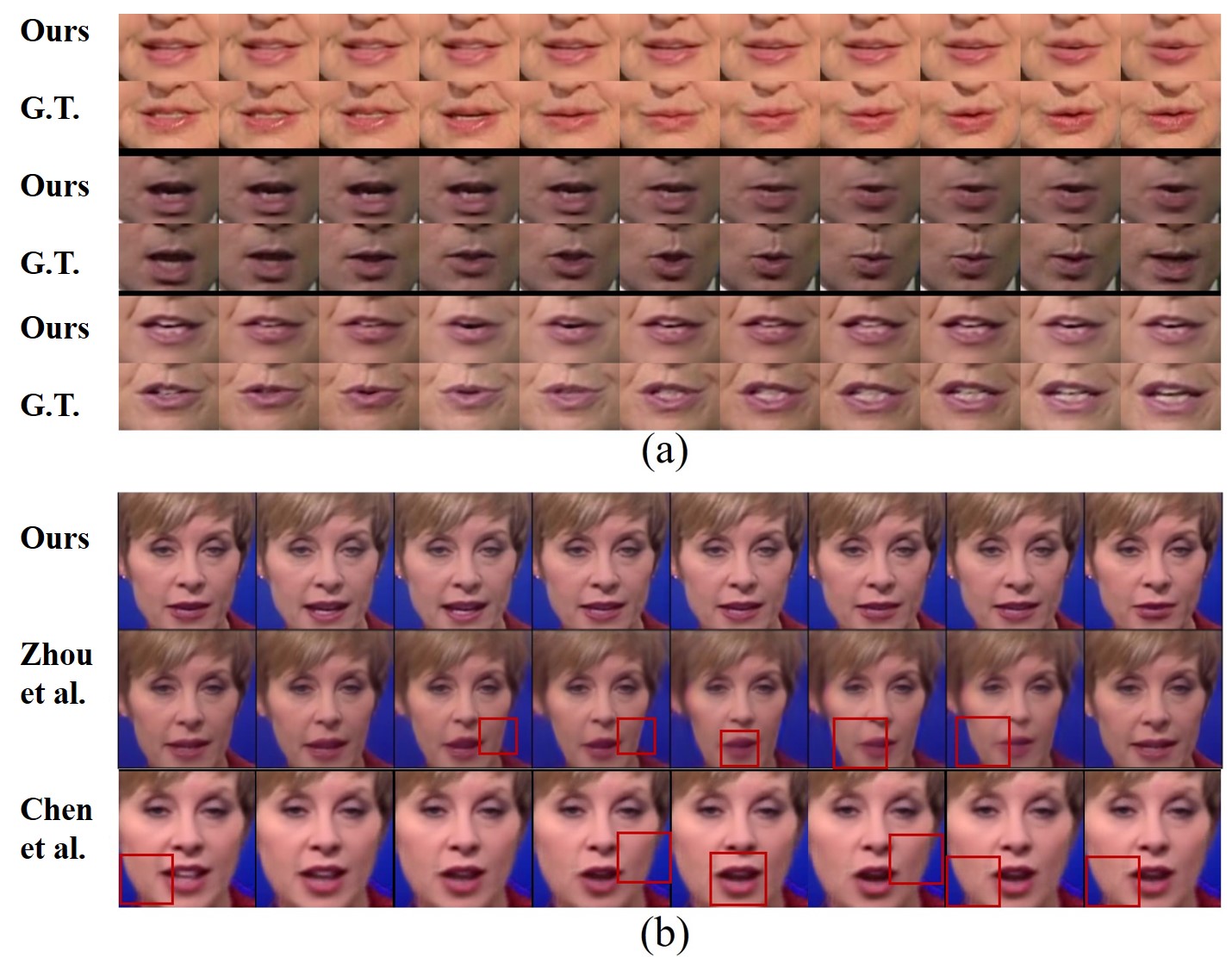}
		\caption{Generation examples of our method comparing with Ground Truth (G.T.) (a), and Zhou \textit{et al.} and Chen \textit{et al.} (b). (Better zoom in to see the detail).  
		}
		\label{fig:compare_results}
	\end{figure}

	\subsection{Quantitative Results}
	
	We compare our model with four recent state-of-the-art methods, including Chung \textit{et al.} \shortcite{Chung2017YouST}, Chen \textit{et al.} \shortcite{Chen2018LipMG} Zhou \textit{et al.} \shortcite{Zhou2018TalkingFG}, and Chen \textit{et al.} \shortcite{chen2019hierarchical}. 
	Table \ref{tbl:comparsion-results} shows the quantitative results of our method and its competitors with higher PSNR, SSIM and lower LMD, suggesting the best quality of the generated video frames of the talking faces.
	Although Zhou \textit{et al.} \cite{Zhou2018TalkingFG} obtains the lowest PSNR, it obtains the second highest SSIM and its SSIM is significantly better than Chung \textit{et al.} \cite{Chung2017YouST} and Chen \textit{et al.} \cite{Chen2018LipMG}. 
	
	To further verify the robustness for arbitrary person generation, we evaluate our method on another benchmark dataset GRID \cite{cooke2006audio} and report the comparison results in Table \ref{tbl:crossdataset-comparsion-results}.
	From Table \ref{tbl:crossdataset-comparsion-results}, we observe that our model achieves the highest SSIM and the lowest LMD, demonstrating the effectiveness and robustness of our method. 
	Although the PSNR of our method is a little lower than that of Chen \textit{et al.} \cite{chen2019hierarchical}, our method surpass \cite{chen2019hierarchical} on the metrics of both SSIM and LMD.
	Therefore, our method always achieves the highest PSNR and SSIM, demonstrating our method is able to generate high-quality videos.

	\begin{table}[t]
		\centering

		\begin{tabular}{l|lllll} 
			\toprule
			\cline{1-6} 
			\multirow{2}{*}{Methods} & \multicolumn{4}{c}{Evaluation on LRW}   \\
			\cline{2-6}  
			&    PSNR $\uparrow$&  SSIM $\uparrow$& LMD $\downarrow$\\
			\midrule
			Chung  \textit{et al.} \shortcite{Chung2017YouST} &  28.06 & 0.46  &  2.23 \\\hline
			Chen  \textit{et al.} \shortcite{Chen2018LipMG} &  28.65 & 0.53  & 1.92 \\\hline
			Zhou \textit{et al.} \shortcite{Zhou2018TalkingFG} &  26.80 & 0.88 & --- \\\hline
			Chen \textit{et al.} \shortcite{chen2019hierarchical} &  30.91 & 0.81 & 1.37 \\\hline\hline
			AMIE (Ours) &  \textbf{32.08} & \textbf{0.92}& \textbf{1.21}\\\hline
		\end{tabular}
		\caption{Quantitative results. }
		\label{tbl:comparsion-results}
		
	\end{table}
	
	\begin{table}[t]
		\centering

		\begin{tabular}{l|lllll} 
			\toprule
			\cline{1-6} 
			\multirow{2}{*}{Methods} & \multicolumn{4}{c}{Evaluation on GRID}   \\
			\cline{2-6}  
			&    PSNR $\uparrow$&  SSIM $\uparrow$& LMD $\downarrow$\\
			\midrule
			Chung \textit{et al.} \shortcite{Chung2017YouST} &  29.36 & 0.74 &   1.35 \\\hline
			Chen \textit{et al.} \shortcite{Chen2018LipMG} &  29.89 & 0.73  & 1.18 \\\hline
			Chen \textit{et al.} \shortcite{chen2019hierarchical} &  \textbf{32.15} & 0.83  & 1.29 \\\hline\hline
			AMIE (Ours) &  31.01 & \textbf{0.97}&  \textbf{0.78}\\\hline
		\end{tabular}
		\caption{Cross-dataset evaluation of our method on GRID dataset pre-trained on LRW dataset.}
		\label{tbl:crossdataset-comparsion-results}
	\end{table}

	\begin{figure}[tb]
		\centering
		\includegraphics[width=0.82\linewidth]{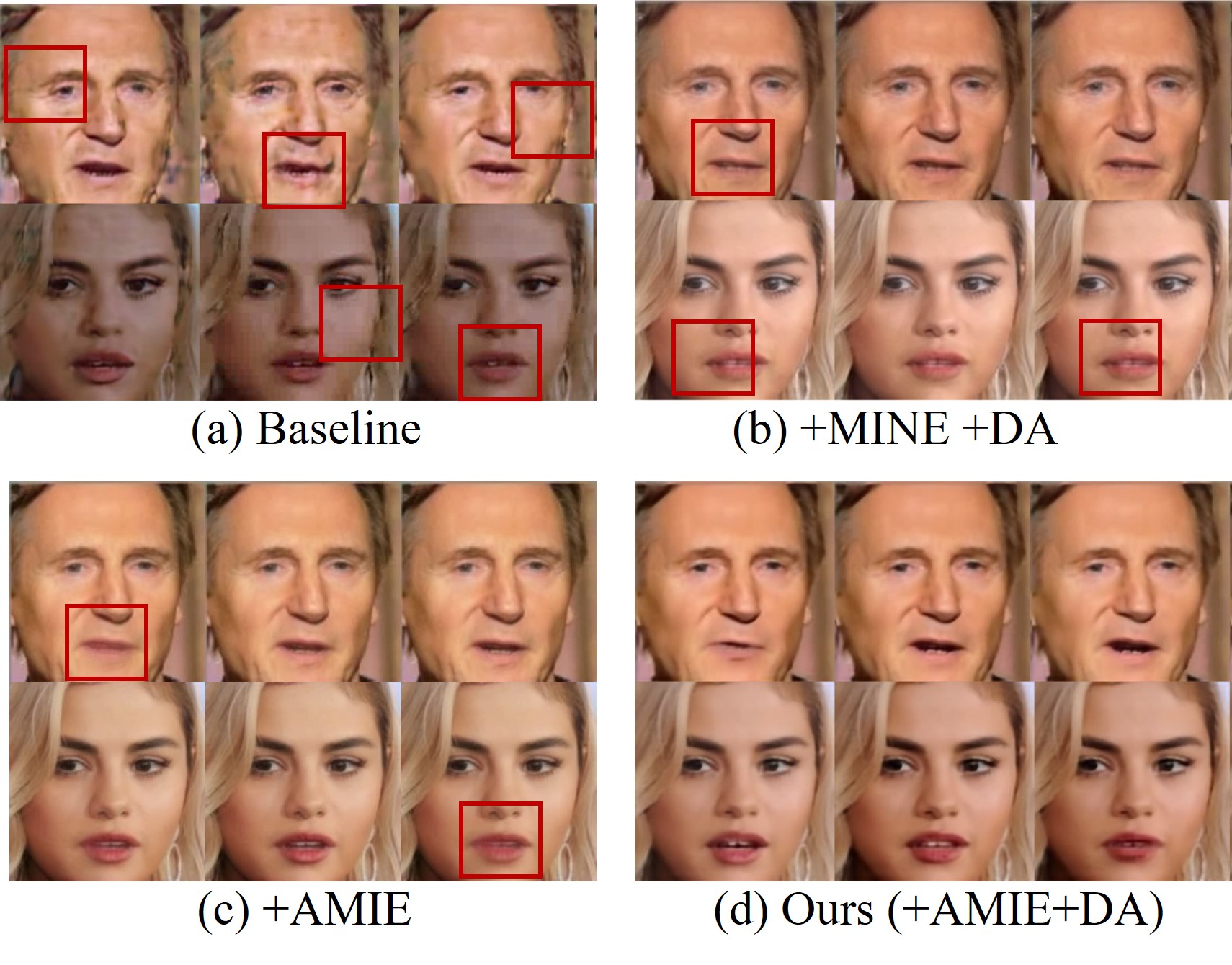}
		\caption{Qualitative results of ablation. }
		\label{fig:abl_compare_results}
	\end{figure}

	\subsection{Qualitative Results}
	To present the superiority of our method, we provide generated samples compared with Zhou \textit{et al.} \shortcite{Zhou2018TalkingFG}, and Chen \textit{et al.} \shortcite{chen2019hierarchical}. 
	As shown in Fig. \ref{fig:compare_results} (a), the identity face image is obtained from the testing set in LRW \cite{Chung16} dataset, while in Fig. \ref{fig:compare_results} (b), the arbitrary identity face image is downloaded from internet. 
	It is clear to see that, Zhou \textit{et al.} suffer from a "zoom-in-and-out" effect, while the lip shapes of Chen \textit{et al.} appear differences from the real one. 
	In general, our model can generate more realistic and synchronous frames.

	\begin{table}[t]
		\centering

		\begin{tabular}{l|lllll} 
			\toprule
			\cline{1-6} 
			\multirow{2}{*}{Methods} & \multicolumn{4}{c}{Evaluation on LRW}   \\
			\cline{2-6}  
			&    PSNR $\uparrow$ &  SSIM $\uparrow$ & LMD $\downarrow$\\
			\midrule
			Baseline &  28.88 & 0.89 & 1.36 \\\hline
			(a) + MINE &  29.05 & 0.89 & 1.38 \\\hline
			(b) + DA &  29.19 & 0.90 & 1.37 \\\hline
			(c) + MINE + DA  &  29.08 & 0.89  & 1.32 \\\hline
			(d) + MINE + JS  & 29.33  & 0.90  & 1.50 \\\hline
			\hline
			(e) + MINE + JS + DA &  29.12 & 0.91  & 1.21 \\\hline
			(f) + MINE + $Asy.$ + DA  &  29.30 & 0.90  & 1.33 \\\hline
			(g)  + AMIE   &  29.41 & \textbf{0.92}  & 1.22 \\\hline
			\hline
			Ours (+ AMIE + DA) 
			&  \textbf{29.64} & \textbf{0.92}&  \textbf{1.18}\\\hline
			
		\end{tabular}
		
		\caption{Ablation study of the key components AMIE and DA in our method as well as two strategies applied in AMIE: Asymmetric Training ($Asy.$) and JS represented estimator (JS). Ours = Baseline + AMIE + DA, and AMIE = MINE + $Asy.$ + JS. }
			
		\label{tbl:ablation-study}
		
	\end{table}
	
	\subsection{Ablation Study}
	In order to quantify the effect of each component of our method, we conduct ablation study experiments to verify the contributions of the two key components in our model: Asymmetric Mutual Information Estimator (AMIE) and Dynamic Attention (DA), and the two important strategies in AMIE: the asymmetric training strategy ($Asy.$) and JS represented estimator. 
	From Table \ref{tbl:ablation-study}, it is clear to see that,
	(1) Simply adopting MI into our baseline did not significantly improve the results, comparing Table \ref{tbl:ablation-study} (a) to baseline.
	(2) DA plays an important role in talking face generation, comparing Table \ref{tbl:ablation-study} (b) to baseline.
	(3) Introducing either $Asy.$ or JS strategy shows the great improvement on almost metrics, comparing Table \ref{tbl:ablation-study} (e) and (f) to Table \ref{tbl:ablation-study} (c).
	(4) After integrating the asymmetric training strategy ($Asy.$) and JS represented estimator into the MINE, our AMIE can further improve the performance on all the metrics, as shown in Table \ref{tbl:ablation-study} (g).
	(5) Integrating AMIE (comparing Table \ref{tbl:ablation-study} 'Ours' to Table (b)) can further boost the performance, while simply integrating MINE (comparing Table \ref{tbl:ablation-study} (c) to (b)) the performance declined, that verifies the contribution of proposed AMIE. 
	Fig. \ref{fig:abl_compare_results} demonstrates the visualized examples.
	
	\subsection{User Study}
	We conduct a user study on the LRW dataset with 42 volunteers in both realistic and synchronization of generation. 
	We randomly select the samples generated by our method, Zhou \textit{et al.} \shortcite{Zhou2018TalkingFG} and Chen \textit{et al.} \shortcite{chen2019hierarchical}. 
	Then, the volunteers were asked to answer the following two questions: (1) which one appears more realistic and (2) which one provides more temporal synchronism referring to the ground truth. 
	Table \ref{tbl:userstudy} clearly demonstrates that our model achieves the highest rating in both realistic and synchronization. 
	
	\begin{table}[t]
		\centering

		\begin{tabular}{l|cc} 
			\toprule
			\cline{1-3} 
			Methods    &    Realistic &  Synchronization \\
			\midrule
			Zhou \textit{et al.} \shortcite{Zhou2018TalkingFG}&  15.22\%& 18.17\% \\\hline
			Chen \textit{et al.} \shortcite{chen2019hierarchical} &  28.37\%& 32.92\% \\\hline\hline
			AMIE (Ours) &  \textbf{56.41\%}& \textbf{48.91\%}\\\hline
		\end{tabular}

		\caption{Results of user study.}
		\label{tbl:userstudy}
		\end{table}
	
	\section{Conclusion}
	\label{sec:conclusion}
	We have proposed a novel model of talking face generation for arbitrary identities via exploring the cross-modality coherence in this paper. 
	Our model leverages the asymmetric mutual information estimator to learn the correlation of audio and facial image features and utilizes dynamic attention to simulate the process of disentangling. Extensive experimental results on benchmark datasets demonstrate the promising performance of our method.
	\section*{Acknowledgments}
	This work is partially funded by the Beijing Natural Science Foundation (JQ18017), Youth Innovation Promotion Association CAS (Grant No. Y201929), the National Nature Science Foundation of China (61976002), and the Open Project Program of the National Laboratory of Pattern Recognition (NLPR) (201900046). 
		\bibliographystyle{named}
		\bibliography{refs_mini}

\end{document}